\documentclass{article}

\usepackage{microtype}
\usepackage{graphicx}
\usepackage{subfigure}
\usepackage{caption}
\usepackage{booktabs} 
\usepackage{color,soul}
\usepackage{bm}
\usepackage{tabularx}

\usepackage{hyperref}

\usepackage[accepted]{icml2019}

\icmltitlerunning{Recovering the parameters underlying the Lorenz-96 chaotic dynamics}

\begin{document}

\twocolumn[
\icmltitle{Recovering the parameters underlying the Lorenz-96 chaotic dynamics}



\icmlsetsymbol{equal}{*}

\begin{icmlauthorlist}
\icmlauthor{Soukayna Mouatadid}{to}
\icmlauthor{Pierre Gentine}{goo}
\icmlauthor{Wei Yu}{to}
\icmlauthor{Steve Easterbrook}{to}
\end{icmlauthorlist}

\icmlaffiliation{to}{Department of computer science, University of Toronto, ON, Canada}
\icmlaffiliation{goo}{Department of Earth and Environmental Engineering, Earth Institute, and Data Science Institute, Columbia University, New York, NY, USA}

\icmlcorrespondingauthor{Soukayna Mouatadid}{soukayna@cs.toronto.edu}

\icmlkeywords{Climate Modelling, Chaos, Deep Learning, ICML}

\vskip 0.3in
]



\printAffiliationsAndNotice{}  

\begin{abstract}
Climate projections suffer from uncertain equilibrium climate sensitivity. The reason behind this uncertainty is the resolution of global climate models, which is too coarse to resolve key processes such as clouds and convection. These processes are approximated using heuristics in a process called parameterization. The selection of these parameters can be subjective, leading to significant uncertainties in the way clouds are represented in global climate models. Here, we explore three deep network algorithms to infer these parameters in an objective and data-driven way. We compare the performance of a fully-connected network, a one-dimensional and, a two-dimensional convolutional networks to recover the underlying parameters of the Lorenz-96 model, a non-linear dynamical system that has similar behavior to the climate system.
\end{abstract}

\section{Introduction}

The global warming target of the 2015 Paris Agreement is $2^\circ$C above pre-industrial level. How much $CO_2$ can accumulate in the atmosphere before this threshold is crossed? There is no certain answer to this question. There has not been one for decades. The answers vary from 480 ppm, which will be reached around 2030, to 600 ppm, which will be reached much later after 2060 \cite{schneider2017climate}. Optimal emission pathways, policy responses, and socioeconomic costs of climate change vary vastly between the high and low end of this range. 

What lies behind this recalcitrant uncertainty is the coarse resolution of global climate models (GCMs), which hinders resolving crucial processes like cloud formation and moist convection  \cite{stevens2013climate}. Climate models compute solutions to the laws of thermodynamics and fluid dynamics on computational grids. These grids have a typical scale varying from 10 to 150 km. Cloud formation takes place on smaller scales (i.e., ~2 km or less) and cannot be resolved by current climate models. Therefore, clouds are modeled by heuristically approximated parameterization schemes (i.e., rule-of-thumb principles). These parameterization schemes include uncertain parameters, selected by climate scientists based on experience and intuition \cite{hourdin2017art}. In addition, the non-linearity of the climate system means that climate simulations can depend sensitively and in unexpected ways on these parameters \cite{zhao2016uncertainty}, which is why we still cannot answer our initial question with certainty. In this study, we show that deep learning can provide an objective, data-driven and computationally efficient approach to parameters estimation for sub-grid parameterization. We illustrate this idea using the Lorenz-96 model \cite{lorenz1996predictability}.

\section{Lorenz-96 model}

The Lorenz-96 model is a dynamical system that behaves in a non-linear way resembling the non-linear behavior of the climate system and consists of a coupling of variables evolving over slow and fast timescales \cite{schneider2017earth}. The model consists of $K$ slow variables, where each slow variable $X_k, k = 1,...,K$ is coupled with $J$ fast variables $Y_{j,k}, j = 1,...,J$. The model is governed by $K + JK$ equations:

\begin{equation}
\frac{dX_k}{dt} = -X_{k-1}(X_{k-2}-X_{k+1})-X-k +F-hc\overline{Y}_k\\
\end{equation}
\begin{equation}
\frac{1}{c}\frac{dY_{j,k}}{dt} = -bY_{j+1,k}(Y_{j+2,k}-Y_{j-1,k})-Y_{j,k}+\frac{h}{J}X_k\\
\end{equation}
where
\begin{equation}
\overline{Y}_k = \frac{1}{J} \sum_{j=1}^{J} Y_{j,k}\\
\end{equation}

The slow variables $\bm{X}$ represent resolved-scale variables in a climate model, while the fast variables $\bm{Y}$ represent unresolved variables (e.g., cloud convection). This set of equations are coupled through the mean term $\overline{Y}_k$ and this coupling is controlled by three keys parameters: $b$, $c$ and $h$. The parameter $b$ controls the amplitude of the nonlinear interactions among the fast variables, while the parameter $c$ controls how rapidly the fast variables fluctuate relative to the slow variables and the parameter $h$ controls how strong the coupling between the fast and slow variables is. The goal of this study is to investigate the viability of parameter estimation using three learning models, namely a fully connected network (FC), a one-dimensional convolutional network (Conv1D) and a two-dimensional convolutional network (Conv2D).  

\section{Data and methodology}

The Lorenz-96 (L-96) model was used to generate 200 simulations for four slow variables X, each associated with four fast variables Y. Each simulation was generated using a different combination of $b$, $c$ and $h$ parameters and was initialized with the same values for X and Y. The resulting temporal sequences were mapped to grayscale images of size (50000, 20), where the width dimension represents the slow and fast variables (i.e., 4+16) and the height dimension represents the number of time steps over which the L-96 was accumulated. Each image was split into 2500 smaller (20, 20) image chunks. These chunks were used as training examples for the networks. For all learning models explored in this study, the input to the model is an image chunk of shape (20, 20, 1) and the output layer is an FC layer with three nodes outputting a prediction for $b$, $c$ and h. During testing, the inferred values for the parameters $b$,c and $h$ were averaged over the chunks belonging to the same initial L-96 image. The networks were optimized using Adam.  The loss function used was the weighted mean squared error normalized by the standard deviations of the parameters. 

The experiments in this study were based on using 200 simulations, as this dataset size was large enough to represent different behaviors of the Lorenz-96 model, with sufficiently long and diverse sequences for training. Also, the Lorenz model was accumulated for this long (i.e., 50,000 time steps) to be able to characterize the trajectories of the different parameter combinations. As the model shows chaotic behavior, using a smaller number of time steps might result in trajectories that are initially close to each other before later diverging. A run time of 50,000 ensures that potential diverging trajectories of the variables $X$ and $Y$ have time to unravel and be clearly displayed in the training dataset. In addition, considering that the Lorenz model hardly repeats local trajectory behaviors, using patches of shape (20, 20) can be considered as separate training examples showcasing the possible local behaviors associated with a given parameters combination.

Since the L-96 data is used to represent climate data with hidden parameters, and considering that the majority of climate data is spatio-temporal, training an Conv2D to recover the parameters seems to be an ideal architecture choice. However, in our case, the L-96 images are temporal on the y-dimension while the x-dimension represents the different slow and fast variables. Therefore, while using a convolutional neural network is a more sensitive approach for actual climate data, in the case of the L-96 images, we start by using a fully-connected network, where every image chunk is first flattened and then fed to three dense layers with 400, 200 and 60 nodes.  Additionally, we also train a one dimensional convolutional model along the temporal dimension (y axis of the L-96 image), to investigate whether patterns across time can be learned by the network. In this case, we based the model on two 1D convolutional layer with 32 filters of size 3 in each layer, followed by a maxpooling layer and two dense layers with 128 and 60 nodes. Finally, although being aware that the L-96 data is not spatio-temporal and therefore breaks the assumptions for using a Conv2D model, as the x dimension does not contain any spatial information, we investigate this architecture to test the potential that 2D convolutional networks can hold when trained on actual spatio-temporal climate data. It should be noted here that using a Conv2D assumes that two adjacent columns in a given L-96 image chunk have a stronger correlation than two further apart columns. This might or might not be true, and we expect the Conv2D to fail at capturing long-ranging relationships between pixel columns far from each other (e.g., $Y_1$ and $Y_10$). In all architectures, LeakyRelu was used as the activation function with an alpha equal to 0.001.

Two learning tasks were investigated: first the parameters were recovered from both the slow and fast variables used as inputs, then the parameters were recovered from the fast variables alone. In a second set of experiments, two different testing modes were evaluated. In the first mode, referred to as $test\_mode = False$, the image chunks in the test set were unobserved during training but came from the same initial (50000, 20) L-96 images (i.e., the same parameters space), while in the second scenario, when $test\_mode = True$, the image chunks at test time came from newly generated (50000, 20) L-96 images.

\section{Results}

\begin{figure}[ht]
\begin{center}
\begin{subfigure}
  \centering
  \includegraphics[width=.450\columnwidth]{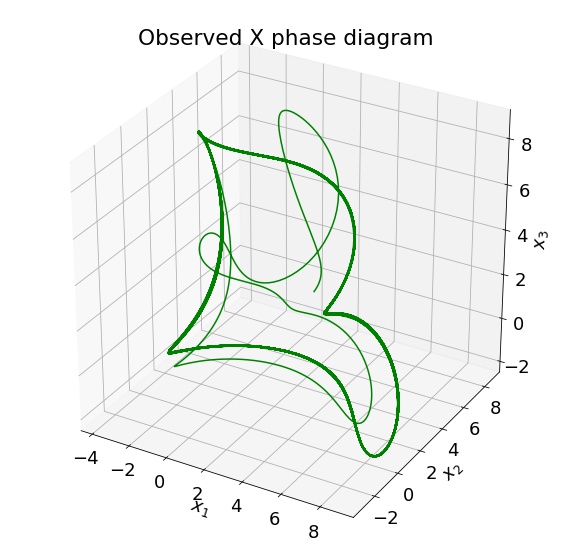}
\end{subfigure}%
\begin{subfigure}
  \centering
  \includegraphics[width=.450\columnwidth]{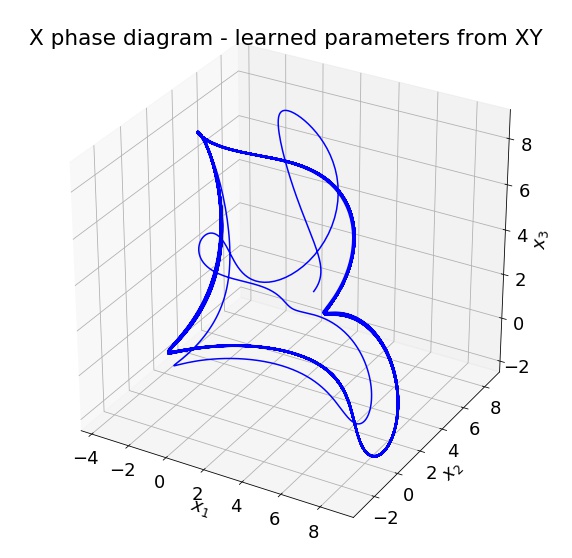}
\end{subfigure}
\begin{subfigure}
  \centering
  \includegraphics[width=.450\columnwidth]{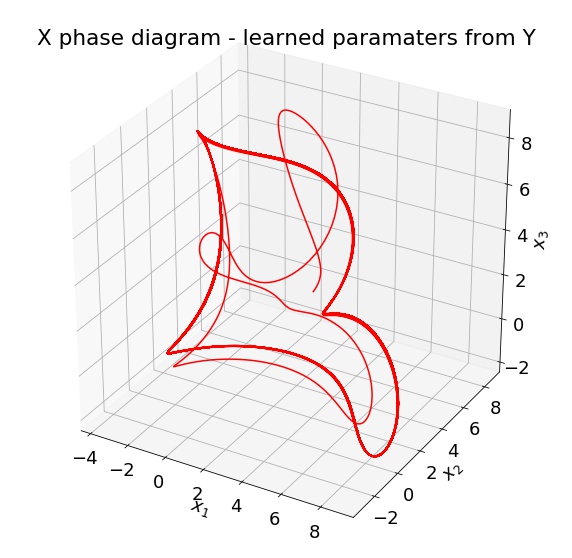}
\end{subfigure}
\begin{subfigure}
  \centering
  \includegraphics[width=.450\columnwidth]{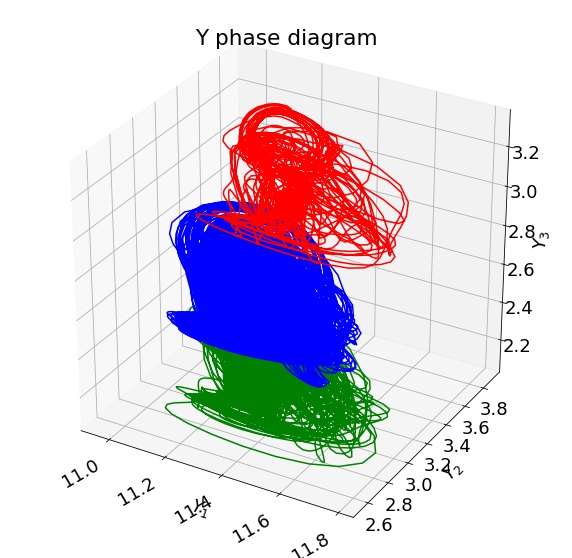}
\end{subfigure}
\caption{Lorenz-96 phase diagram of the first three slow (X) and fast (Y) variables using observed parameters (green), learned parameters from the X and Y variables (blue) and learned parameters from the Y variables only (red). The learning algorithm is a fully connected network with $test\_mode$ set to $False$.}
\label{Figure_1}
\end{center}
\end{figure}

\begin{figure}[ht]
\begin{center}
\begin{subfigure}
  \centering
  \includegraphics[width=.450\columnwidth]{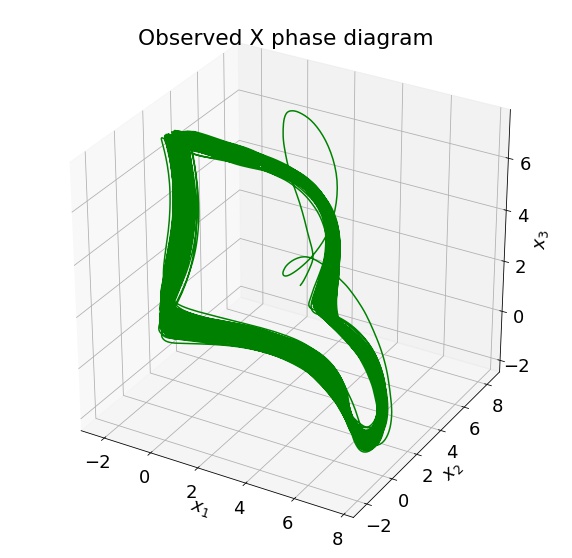}
\end{subfigure}%
\begin{subfigure}
  \centering
  \includegraphics[width=.450\columnwidth]{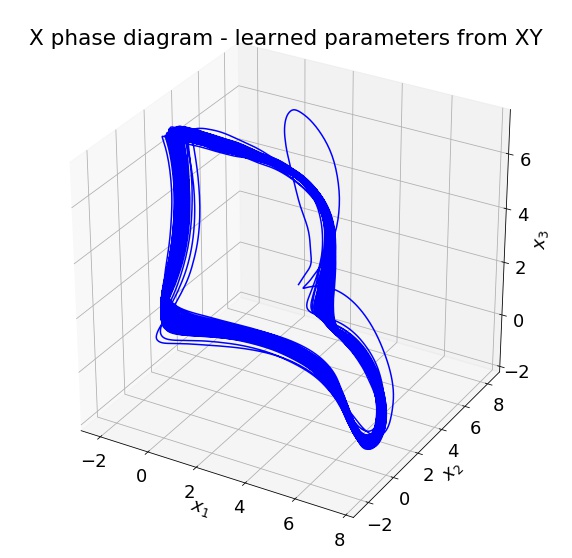}
\end{subfigure}
\begin{subfigure}
  \centering
  \includegraphics[width=.450\columnwidth]{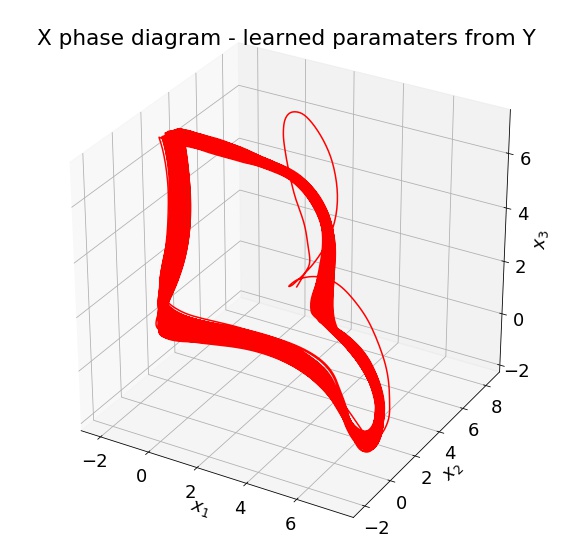}
\end{subfigure}
\begin{subfigure}
  \centering
  \includegraphics[width=.450\columnwidth]{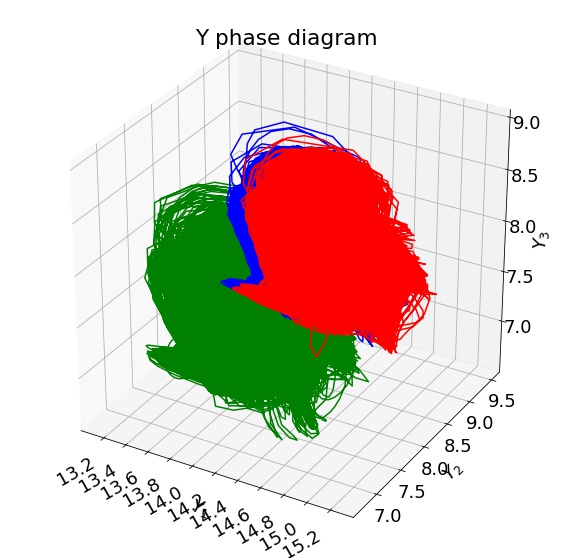}
\end{subfigure}
\caption{Lorenz-96 phase diagram of the first three slow (X) and fast (Y) variables using observed parameters (green), learned parameters from the X and Y variables (blue) and learned parameters from the Y variables only (red). The learning algorithm is a 1D convolutional model with $test\_mode$ set to $True$.}
\label{Figure_2}
\end{center}
\end{figure}

\begin{figure}[ht]
\begin{center}
\begin{subfigure}
  \centering
  \includegraphics[width=.47\columnwidth]{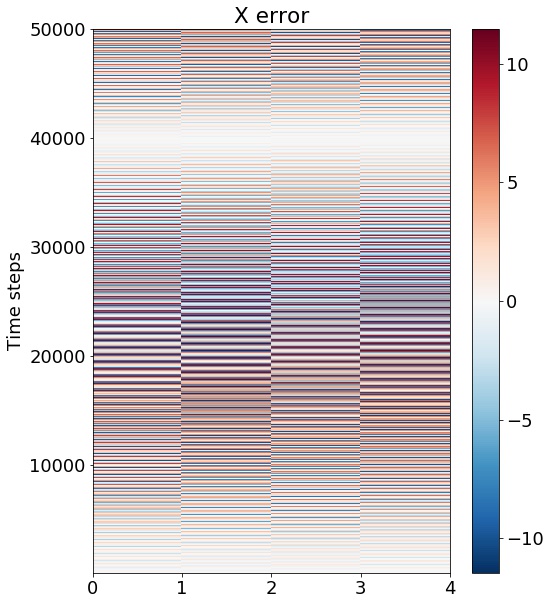}
\end{subfigure}%
\begin{subfigure}
  \centering
  \includegraphics[width=.4\columnwidth]{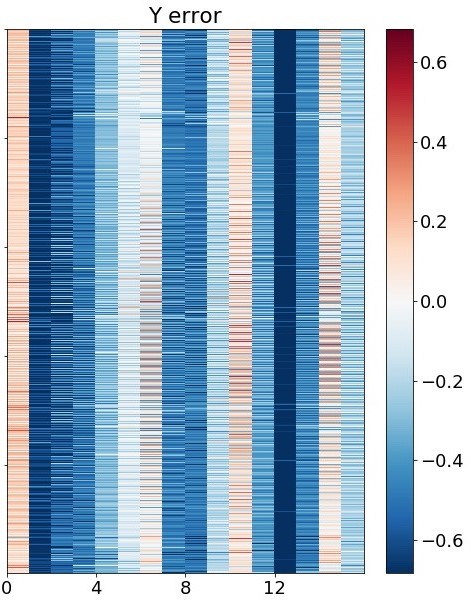}
\end{subfigure}
\begin{subfigure}
  \centering
  \includegraphics[width=.47\columnwidth]{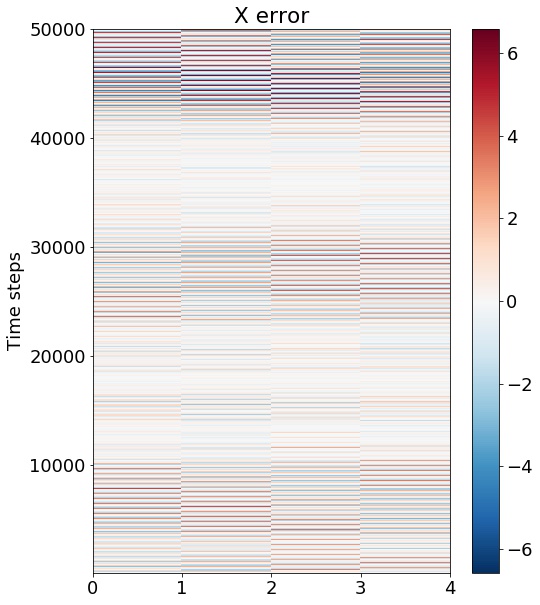}
\end{subfigure}
\begin{subfigure}
  \centering
  \includegraphics[width=.4\columnwidth]{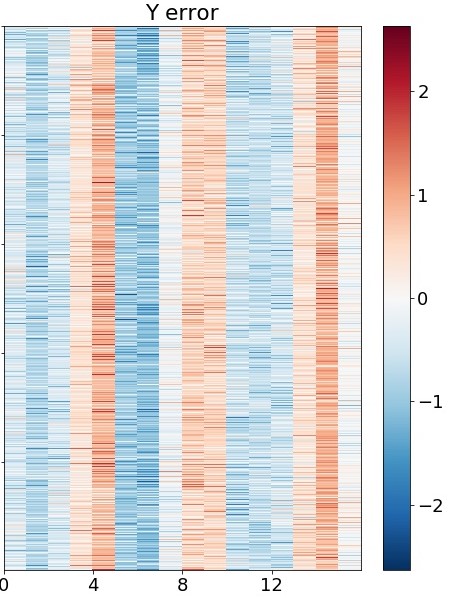}
\end{subfigure}
\caption{Errors between the Lorenz-96 slow (X) and fast (Y) variables generated using the observed parameters and the inferred parameters using the FC model trained on the Y variables with $test\_mode$ set to $False$ (top row) and using the Conv1D model trained on Y variables with $test\_mode$ set to $True$ (bottom row).}
\label{Figure_3}
\end{center}
\end{figure}

Overall, all of the FC, Conv1D and Conv2D models successfully recovered the parameters $b$, $c$ and $h$, as measured by the MSE loss and the coefficient of determination $r^2$ (see Table~\ref{Table_1}). In addition, the performance of linear models (LR) is shown as a baseline. Qualitative results are presented in Figure~\ref{Figure_1},~\ref{Figure_2} and~\ref{Figure_3}. Using fast variables alone as inputs was sufficient to accurately recover the hidden parameters. Regarding the different test modes explored, when $test\_mode$ was set to $True$, the models’ $r^2$ values dropped by around 0.1, as both the temporal variability (where does the image chunk belong in the original L-96 sequence) and the parameters variability (which combination of $b$,c and $h$ generated a given image chunk) had to be captured by the learning models. 

\begin{table}[t]
\caption{Training and testing loss and coefficient of determination.}
\begin{center}
\begin{small}
\begin{sc}
\resizebox{\columnwidth}{!}
{
\begin{tabular}{llcccc}
\toprule
test mode & model &train loss & test loss & train $r^2$ & test $r^2$ \\
\midrule
\textcolor{white}{.} &\textcolor{white}{.} &\multicolumn{4}{c}{Learning from X and Y}\\ 
\cmidrule{3-6}
\textcolor{white}{.}   & LR &1.7512 &1.7560	&0.7588 &0.7578 \\
\textcolor{white}{.}   & FC    & 0.6583 &0.6714 &0.9094 &0.9074  \\
\textcolor{white}{.}    & Conv1D& 0.6682 &0.6812 &0.9079 &0.9060 \\
\textcolor{white}{.}   & Conv2D& 0.6502 &0.6861 &0.9105 &0.9054 \\
\cmidrule{3-6}
$False$  & \textcolor{white}{.}  & \multicolumn{4}{c}{Learning from Y only}\\
\cmidrule{3-6}
\textcolor{white}{.}   & LR &1.7394 &1.7429 &0.7605 &0.7597 \\
\textcolor{white}{.}   & FC  &0.6647 &0.6808 &0.9084 &0.9061 \\
\textcolor{white}{.}   & Conv1D &0.6968 &0.7073 &0.9041 &0.9024 \\
\textcolor{white}{.}   & Conv2D &0.6744 &0.7063 &0.9071 &0.9026\\
\midrule
\textcolor{white}{.} &\textcolor{white}{.} &\multicolumn{4}{c}{Learning from X and Y}\\ 
\cmidrule{3-6}
\textcolor{white}{.}   & LR &1.7371 &2.9112 &0.7609 &0.6059 \\
\textcolor{white}{.}   & FC    & 0.7064 &1.3262 &0.9028 &0.8212 \\
\textcolor{white}{.}    & Conv1D& 0.7029 &1.2822 &0.9031 &0.8263 \\
\textcolor{white}{.}   & Conv2D& 0.6577 &1.3260 &0.9070 &0.8125 \\
\cmidrule{3-6}
$True$  & \textcolor{white}{.}  & \multicolumn{4}{c}{Learning from Y only}\\
\cmidrule{3-6}
\textcolor{white}{.}   &LR &1.7407 &2.9268 &0.7604 &0.6039 \\
\textcolor{white}{.}   & FC  &0.6805 &1.3197 &0.9063 &0.8220 \\
\textcolor{white}{.}   & Conv1D &0.6898 &1.2726 &0.9050 &0.8276 \\
\textcolor{white}{.}   & Conv2D &0.6577 &1.3260 &0.9094 &0.8210 \\
\bottomrule
\end{tabular}
}
\end{sc}
\end{small}
\end{center}
\label{Table_1}
\end{table}

In addition, although the FC model lead to the most accurate recovery of parameters when $test\_mode$ was set to $False$, Conv1D outperformed the FC model when $test\_mode$ was changed to $True$. We explain such inconsistency from the perspective of finite difference analysis and optimization. Since the filter size of convolutions is set to 3, the convolutional filters applied on the temporal dimension can capture at most the local second-order relationships in the system while fully-connected counterpart can estimate the higher-order relationships. However, the latter is harder to optimize due to its large number of parameters and less inductive bias. Inductive bias is known to be crucial to deal with unseen situations as it is the case when $test\_mode= True$. As for the  $test\_mode= False$ setting, the fully-connected model can depict more complex temporal dependency than convolutional filter and thus extract more useful representations for parameter recovery.

In order to illustrate how the model proposed in this paper fits into a GCM, we adopt the framework proposed in~\cite{schneider2017climate}.  In~\cite{schneider2017climate}, the authors make the distinction between two types of parameters: computable and non-computable parameters.  Computable parameters can, in principle, be inferred from high-resolution simulations. Non-computable parameters are parameters that, currently, cannot be inferred from high-resolution simulations, either because of computational limitations or because the microscopic equations governing certain processes are unknown. Such non-computable parameters can be learned from global observational records like satellite data, space-based measurements of biogeochemical tracers, etc. The authors further envision a parameterization scheme, which once embedded in a GCM, can learn directly from global observations, with targeted high-resolution simulations used to update parameters in grid cells with the highest uncertainty. In this paper, we illustrate an intermediate step where the parameters are learned off-line, using the Lorenz-96 model. Our first learning task: learning from both slow large-scale variables $X$ and fast small-scale variables $Y$ corresponds to learning parameters from global observations. In this case, the dynamical system (i.e., Lorenz-96) with parameters $b$, $c$ and $h$ represents the GCM, while the $X$ and $Y$ data generated by the dynamical system with the true values of $b$, $c$ and $h$ represent the global observations. The second learning task consists of learning parameters from the fast small-scale $Y$ variables alone and is equivalent to learning about computable parameters from high-resolution simulations.

\section{Conclusion}

Since its introduction in the early 1960s, a wide range of solutions have been proposed for the cumulus parameterization problem \cite{arakawa2004cumulus}. The more recent solutions propose learning models trained on data from different cloud resolving model simulations. Knowing that this data is imperfect and contributes to the cumulus parameterization uncertainty problem, it is likely that the proposed models will end up learning the inaccuracies and structural uncertainties that plague sub-grid processes such as cloud formation. Addressing this issue means learning directly from observations. Using the L-96 model as ground-truth, this study showed the promising potential of deep learning algorithms for objective and data-driven parameters estimation.


\bibliographystyle{icml2019}

\begin{thebibliography}{7}
\providecommand{\natexlab}[1]{#1}
\providecommand{\url}[1]{\texttt{#1}}
\expandafter\ifx\csname urlstyle\endcsname\relax
  \providecommand{\doi}[1]{doi: #1}\else
  \providecommand{\doi}{doi: \begingroup \urlstyle{rm}\Url}\fi

\bibitem[Arakawa(2004)]{arakawa2004cumulus}
Arakawa, A.
\newblock The cumulus parameterization problem: Past, present, and future.
\newblock \emph{Journal of Climate}, 17\penalty0 (13):\penalty0 2493--2525,
  2004.

\bibitem[Hourdin et~al.(2017)Hourdin, Mauritsen, Gettelman, Golaz, Balaji,
  Duan, Folini, Ji, Klocke, Qian, et~al.]{hourdin2017art}
Hourdin, F., Mauritsen, T., Gettelman, A., Golaz, J.-C., Balaji, V., Duan, Q.,
  Folini, D., Ji, D., Klocke, D., Qian, Y., et~al.
\newblock The art and science of climate model tuning.
\newblock \emph{Bulletin of the American Meteorological Society}, 98\penalty0
  (3):\penalty0 589--602, 2017.

\bibitem[Lorenz(1996)]{lorenz1996predictability}
Lorenz, E.~N.
\newblock Predictability: A problem partly solved.
\newblock In \emph{Proc. Seminar on predictability}, volume~1, 1996.

\bibitem[Schneider et~al.(2017{\natexlab{a}})Schneider, Lan, Stuart, and
  Teixeira]{schneider2017earth}
Schneider, T., Lan, S., Stuart, A., and Teixeira, J.
\newblock Earth system modeling 2.0: A blueprint for models that learn from
  observations and targeted high-resolution simulations.
\newblock \emph{Geophysical Research Letters}, 44\penalty0 (24),
  2017{\natexlab{a}}.

\bibitem[Schneider et~al.(2017{\natexlab{b}})Schneider, Teixeira, Bretherton,
  Brient, Pressel, Sch{\"a}r, and Siebesma]{schneider2017climate}
Schneider, T., Teixeira, J., Bretherton, C.~S., Brient, F., Pressel, K.~G.,
  Sch{\"a}r, C., and Siebesma, A.~P.
\newblock Climate goals and computing the future of clouds.
\newblock \emph{Nature Climate Change}, 7\penalty0 (1):\penalty0 3,
  2017{\natexlab{b}}.

\bibitem[Stevens \& Bony(2013)Stevens and Bony]{stevens2013climate}
Stevens, B. and Bony, S.
\newblock What are climate models missing?
\newblock \emph{Science}, 340\penalty0 (6136):\penalty0 1053--1054, 2013.

\bibitem[Zhao et~al.(2016)Zhao, Golaz, Held, Ramaswamy, Lin, Ming, Ginoux,
  Wyman, Donner, Paynter, et~al.]{zhao2016uncertainty}
Zhao, M., Golaz, J.-C., Held, I., Ramaswamy, V., Lin, S.-J., Ming, Y., Ginoux,
  P., Wyman, B., Donner, L., Paynter, D., et~al.
\newblock Uncertainty in model climate sensitivity traced to representations of
  cumulus precipitation microphysics.
\newblock \emph{Journal of Climate}, 29\penalty0 (2):\penalty0 543--560, 2016.

\end{thebibliography}

\end{document}